\documentclass[conference]{IEEEtran}

\IEEEoverridecommandlockouts

\usepackage[numbers,sort&compress]{natbib}
\usepackage{amsmath,amssymb,amsfonts}
\usepackage{algorithmic}
\usepackage{graphicx}
\usepackage{textcomp}
\usepackage{multirow}
\usepackage{xcolor}
\usepackage{url}
\usepackage[bottom]{footmisc}

\makeatletter
\renewcommand*{\footnoterule}{%
  \kern -3pt                
  \hrule width 1in height 0.4pt  
  \kern 2.6pt               
}
\makeatother

\def\BibTeX{{\rm B\kern-.05em{\sc i\kern-.025em b}\kern-.08em
    T\kern-.1667em\lower.7ex\hbox{E}\kern-.125emX}}
\begin{document}
\bstctlcite{IEEEexample:BSTcontrol}
\title{Cross-Temporal Sinhala OCR: Page-Level Adaptation and Diachronic Analysis}


\author{\IEEEauthorblockN{Avisha Dilhara\IEEEauthorrefmark{1}, Nevidu Jayatilleke\IEEEauthorrefmark{2}}
\IEEEauthorblockA{\IEEEauthorrefmark{1}School of Computing, Informatics Institute of Technology, Sri Lanka\\
\texttt{avisha.20220213@iit.ac.lk}}
\IEEEauthorblockA{\IEEEauthorrefmark{2}Department of Computer Science \& Engineering, University of Moratuwa, Sri Lanka\\
\texttt{nevidu.25@cse.mrt.ac.lk}}}


\maketitle

\begin{abstract}
Sinhala is a morphologically rich abugida spoken by roughly 16 million people in Sri Lanka, and to date, there are no publicly available real-world datasets for page-level Sinhala OCR. All previous studies for assessing Sinhala OCR models have used artificially generated data. To bridge the gap, we introduce \texttt{sinhala-ocr-lk-acts-1010}, an annotated dataset of 1,010 page-level images and their transcriptions collected from Sri Lankan Legislative Acts published between 1981-1989 and 2000-2019, split into 707 training examples, 101 validation examples, and 202 testing examples. Three models based on deep learning-based visual language processing, namely DeepSeek-OCR~V1, DeepSeek-OCR~V2, and LightOnOCR-2-1B, are fine-tuned using QLoRA in 8 experiments conducted on consumer and cloud GPUs. LightOnOCR-2-1B is the top performer, achieving a CER of 1.05\% across all test examples, outperforming state-of-the-art open-source OCR models such as Surya-OCR (8.84\%) and Tesseract v5 (10.69\%), as well as commercially available OCR models such as Google Document AI (2.06\%). Our results suggest that LightOnOCR-2-1B outperforms other baselines on real-world OCR tasks and maintains consistent performance across all print periods, even when documents are severely degraded.
\end{abstract}

\begin{IEEEkeywords}
Optical Character Recognition, Sinhala, Low-resource Languages, Diachronic Evaluation
\end{IEEEkeywords}

\section{Introduction}

Optical character recognition (OCR) is a key enabling technology for digitising printed documents and making their content searchable and accessible at scale. For low-resource and complex-script languages, OCR accuracy remains significantly lower than for high-resource languages such as English, as shown by  Jayatilleke and de Silva \cite{jayatilleke_zero-shot_2025}.


Sinhala is the primary official language in Sri Lanka. The script of the Sinhala language is an abugida, with many characters, ligatures, and similarities among characters, making it challenging for OCR engines. Existing research has steadily improved the development of OCR engines using Tesseract that can handle Sinhala characters, as well as the development of parallel corpora from government PDFs \cite{anuradha_deep_2020, vasantharajan_adapting_2022}. Recent benchmarking has even been conducted on both commercially available and open-source OCR engines to compare their performance in zero-shot learning on synthetic images \cite{jayatilleke_zero-shot_2025, purushoth_velayuthan_benchmarking_2025}. However, all existing evaluations are conducted on synthetic images or limited font sets, and no publicly available real-printed Sinhala page-level dataset exists.

Page-level Vision Language Models (VLMs) are compatible with such a challenge by processing the whole image of the page in one forward pass, thus bypassing errors associated with a step-by-step layout detection and segmentation process \cite{li_trocr_2023, lv_kosmos-25_2024}. Adaptation of multi-billion-parameter VLMs is made possible due to parameter-efficient fine-tuning based on the QLoRA technique \cite{dettmers_qlora_2023}, which makes adapting billion-parameter VLMs feasible on consumer hardware, as recently demonstrated for low-resource Indic scripts by Kolavi et al. \cite{kolavi_nayana_2025}.

A further gap is \textit{diachronicity}; none of the Sinhala OCR research studies have quantified the effect of temporal variance on accuracy scores, despite proof that recognition accuracy falls from 87\% for contemporary book scans to 67\% for newspaper articles from the 1980s \cite{anuradha_estimating_2021}, commercial engines also deteriorate with older media.\cite{hegghammer_ocr_2021}.

In this paper, we address both gaps. First, we release \texttt{sinhala-ocr-lk-acts-1010}\footnote{\scriptsize \url{https://huggingface.co/datasets/avishadilhara/sinhala-ocr-lk-acts-1010}}, a dataset of 1,010 manually corrected page-level annotated image-text pairs from Sri Lankan legislative acts (1981 -1989, 2000-2019), split into 707 train, 101 validation, and 202 test pairs, made publicly available. Second, we fine-tune three VLMs DeepSeek-OCR~V1 \cite{wei_deepseek-ocr_2025}, DeepSeek-OCR~V2\cite{wei_deepseek-ocr_2026}, and LightOnOCR-2-1B \cite{taghadouini_lightonocr_2026} across eight LoRa \cite{hu_lora_2021} and QLoRA \cite{dettmers_qlora_2023} experiments, achieving a best CER of 1.05\%, surpassing all open-source baselines and Google Document~AI (2.06\%). Third, we conduct the first diachronic evaluation of page-level Sinhala OCR across three temporal periods (1981-1989, 2000-2009, 2010-2019) \cite{springmann_ocr_2017, hegghammer_ocr_2021,schultze_chronicling_2025}.

\section{Related Work}

\subsection{OCR Architectures and Scene Text Systems}

The development of early deep learning OCR led to sequence-based recognition becoming the paradigm. The CRNN model \cite{shi_end--end_2017} utilised CNN-based feature extraction and bidirectional LSTMs, in conjunction with a CTC decoder, without segmentation. In contrast, attention-based rectification architectures such as ASTER \cite{shi_aster_2019} and RARE \cite{shi_robust_2016} achieved better text recognition on curved and distorted text. Following the introduction of the transformer architecture \cite{vaswani_attention_nodate}, which sparked a revolution in vision tasks by introducing the vision transformer \cite{dosovitskiy_image_2021}, the TrOCR model has achieved recent state-of-the-art results in text recognition \cite{li_trocr_2023}, which surpassed all previous models using an encoder-decoder approach without any external language model support. In the realm of detection, the EAST \cite{zhou_east_2017} and CRAFT \cite{baek_character_2019} advanced single-stage text localisation for arbitrary-shaped text, while PhotoOCR \cite{bissacco_photoocr_2013} demonstrated robust recognition in uncontrolled real-world conditions.

\subsection{Low-Resource, Multilingual, and Sinhala OCR}

OCR for low-resource languages remains challenging due to data absence, script complexity, and limited pre-training coverage \cite{agarwal_concise_2024}. Training on authentic over synthetic data is crucial whenever possible \cite{baek_what_2021}, which is particularly applicable to our project on Sinhala. Kolavi et al. \cite{kolavi_nayana_2025} introduced Nayana, an approach that adapts VLMs for OCR in 10 low-resource Indic languages using the LoRA technique and synthetic data, cutting down CER to less than a third of the baseline model. Hegghammer and Thomas\cite{hegghammer_ocr_2021} found that commercial engines are superior to Tesseract when dealing with noisy texts in non-English scripts. On the specific topic of Sinhala, Vasantharajan et al. \cite{vasantharajan_adapting_2022} fine-tuned Tesseract on more than 20 legacy fonts and reduced the initial CER from 7.61\% to 4.74\%, creating a parallel dataset based on government documents, which is the closest work in terms of dataset type. Jayatilleke and de Silva \cite{jayatilleke_zero-shot_2025} compared the zero-shot performance of six engines on synthetic Sinhala-Tamil data and found that Surya achieved the lowest WER of 2.61\% for Sinhala; however, only clean synthetic images were used for evaluation. Purushoth and Ambegoda \cite{purushoth_velayuthan_benchmarking_2025} benchmarked several open-source document image analysis models for Sinhala, finding that Surya-OCR provided the most balanced and accurate performance over legacy models such as Tesseract. The study by Anuradha et al. \cite{anuradha_estimating_2021} reported over 87\% accuracy on modern book text but only 67\% on late 19th-century newspapers (1870–1890), providing the first empirical evidence of diachronic degradation in Sinhala OCR.

\subsection{Page-Level OCR and Diachronic Evaluation}



Standard OCR pipelines often suffer from error cascading across consecutive stages \cite{vempati_why_2025,yu_benchmarking_2025}; page-level VLMs circumvent this problem by processing an entire page in a single pass. KOSMOS-2.5 \cite{lv_kosmos-25_2024,yu_benchmarking_2025} shows the capacity of multimodal LLMs to understand text-heavy document images as a whole. However, LightOnOCR-2-1B \cite{taghadouini_lightonocr_2026}, and DeepSeek-OCR \cite{wei_deepseek-ocr_2025} have shown that it is possible to achieve similar results even in the case of higher resolution document images by employing efficient vision token embeddings enabled by QLoRA \cite{dettmers_qlora_2023} and LoRa \cite{hu_lora_2021}. 

Diachronic evaluation of a model by tracing its performance across different print eras has been identified as essential for measuring its robustness to typographic changes and media decay \cite{springmann_ocr_2017,schultze_chronicling_2025}. Despite the documented period-based degradation in Sinhala \cite{anuradha_estimating_2021}, all prior Sinhala OCR evaluations have been completely synchronic. This work conducts the first controlled diachronic evaluation for Sinhala, spanning 1981--2019.

\section{Dataset Preparation and Preprocessing}

\subsection{Source Documents}

The dataset is sourced from the \texttt{lk\_legal\_docs} GitHub repository \cite{senaratna_sri_2025}, a multilingual resource of Sri Lankan government documents. Each document folder contains a \texttt{metadata.json} file with fields including
\texttt{doc\_type}, \texttt{date\_str}, \texttt{lang}, and \texttt{url\_pdf} pointing to the PDF on \texttt{documents.gov.lk}. Only Sinhala-language (\texttt{lang: "si"}) documents were used; PDFs were downloaded programmatically via the \texttt{url\_pdf} field. as illustrated in Fig.~\ref{fig:pipeline}.

\subsection{Document Processing Pipeline}

\begin{figure}[t]
  \centering
  \includegraphics[width=0.5\columnwidth]{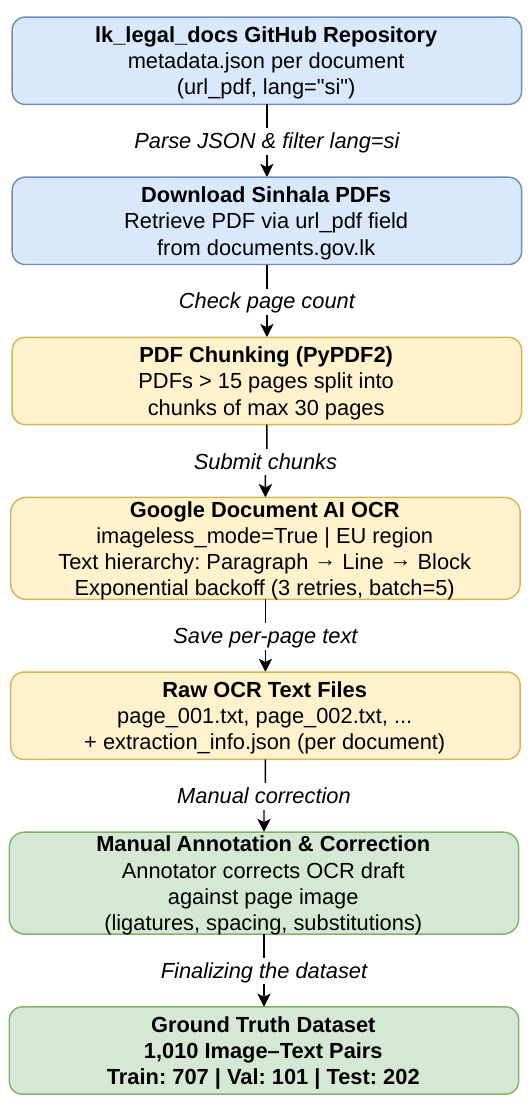}
  \caption{Document processing pipeline for constructing the Sinhala government acts OCR dataset.}
  \label{fig:pipeline}
\end{figure}

PDFs were downloaded programmatically and split into 30-page chunks using PyPDF2 for documents exceeding 15 pages. Each chunk was submitted to Google Document~AI\footnote{\scriptsize \url{https://cloud.google.com/document-ai}} via the \texttt{google-cloud-document-ai} Python client with exponential backoff retry logic (3 retries, 60\,s inter-batch delay); the returned per-page text was saved as UTF-8
\texttt{.txt} files and used as the manual annotation seed, as illustrated in Fig.~\ref{fig:pipeline}.

\subsection{Page Selection and Manual Annotation}

Pages that had well-formed paragraphs in Sinhala were selected for use in our analysis, and other pages, such as forms and tables were not included for the sake of uniformity. Ground-truthing involved manually correcting errors made by Document~AI, including ligature recognition, character substitutions, inconsistent spacing, and erroneous line breaks. Documents from the 1980s needed the most correction, as scans were
of poor quality. This gave us 1,010 data pairs: 410 from the 1980s (1981--1989), 300 from the 2000s (2000--2009), and 300 from the 2010s (2010--2019). The decade 1990--1999 was excluded due to the infeasibility of obtaining complete coverage within the manual annotation timeframe. The selected periods maximise the temporal span of the available diachronic periods.

\subsection{Dataset Splits}

The 1,010 annotated pairs were randomly shuffled with a fixed random seed of 42 to ensure reproducibility, then partitioned into 707 training, 101 validation, and 202 test pairs, with a 70/10/20 ratio. The resulting split maintains balanced representation across all three document eras (1981--1989, 2000--2009, and 2010--2019) in each subset, as documents within each era share similar printing styles and scan characteristics. The final dataset was uploaded to the Hugging Face Dataset Hub as a public repository for version control and reproducibility. Table~\ref{tab:dataset_stats} summarises the dataset statistics.

\begin{table}[t]
\centering
\caption{Sinhala Government Acts OCR Dataset Statistics.}
\label{tab:dataset_stats}
\footnotesize
\resizebox{0.8\columnwidth}{!}{%
\begin{tabular}{lc}
\hline
\textbf{Attribute}            & \textbf{Value}               \\
\hline
Document type                 & Government acts (Sinhala)    \\
Year range                    & 1981--1989, 2000--2019       \\
Total annotated pages         & 1,010                        \\
\quad 1981--1989              & 410                          \\
\quad 2000--2009              & 300                          \\
\quad 2010--2019              & 300                          \\
Training pairs                & 707 (70\%)                   \\
Validation pairs              & 101 (10\%)                   \\
Test pairs                    & 202 (20\%)                   \\
\hline
\end{tabular}}
\end{table}

\begin{figure}[t]
  \centering
  \includegraphics[width=\columnwidth]{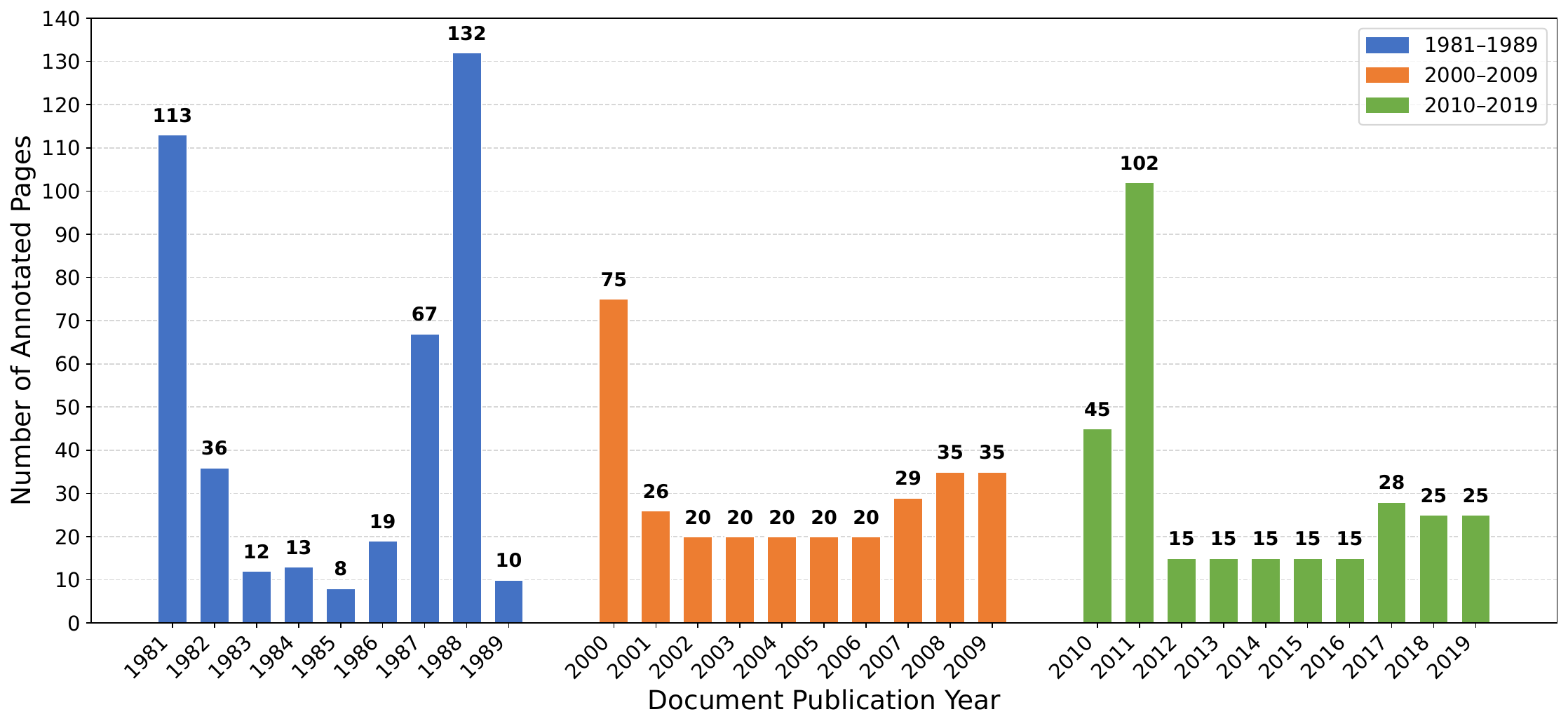}
  \caption{Distribution of annotated page samples across document
  publication years (1981--1989 and 2000--2019).}
  \label{fig:year_dist}
\end{figure}

\section{Experimental Setup}

\subsection{Model Selection}

Three VLMs were selected for their purpose-built design for dense document OCR, script-agnostic recognition, and compatibility with parameter-efficient fine-tuning on available GPU hardware.

\textbf{DeepSeek-OCR~V1}~\cite{wei_deepseek-ocr_2025} proposes the paradigm of context optical compression. The encoder consists of an 80M-parameter SAM-base backbone augmented with window attention, and a 300M-parameter CLIP-large backbone with 16$\times$ convolutional compression, followed by a DeepSeek-3B MoE decoder that uses 570M parameters at each forward propagation step. All experiments used \textit{Gundam} mode (base size 1024\,px, crop size 640\,px) for multi-scale document processing. DeepSeek-OCR~V1 explicitly lists Sinhala among its 100 supported languages.

\textbf{DeepSeek-OCR~V2} \cite{wei_deepseek-ocr_2026} proposes the DeepEncoder~V2, which replaces the CLIP model with a  Qwen2-0.5B LLM-style architecture and introduces causal flow queries that semantically reorder visual tokens before decoding. The local crop size is upgraded from 640\,px to 768\,px in Gundam mode, enabling finer character discrimination.

\textbf{LightOnOCR-2-1B} \cite{taghadouini_lightonocr_2026} is a compact 1.005B-parameter fully-differentiable VLM consisting of a native-resolution ViT encoder pre-trained with Pixtral, a 4$\times$ downsampling multimodal projector, and a Qwen3 language model decoder. Its significantly lower parameter count makes it well-suited to constrained GPU environments, while still supporting complex document layouts, including tables and mathematical formulae.

\subsection{Fine-Tuning with LoRa and QLoRA}

All models were finetuned using LoRa and QLoRA \cite{dettmers_qlora_2023,hu_lora_2021} through Unsloth\footnote{\scriptsize \url{https://unsloth.ai/docs}} for DeepSeek models and through LightOn for LightOnOCR-2-1B. QLoRA maintains the 4-bit quantised NF4 parameters of the base model while adding trainable low-rank adaptation matrices to the attention and feed-forward projections of the transformer decoder. Gradient checkpoint was enabled for all experiments to reduce peak VRAM usage. Table~\ref{tab:experiments} summarises all 8 experiments, which varied GPU hardware, quantisation level, LoRA rank, and input resolution across the three model families, using the same 707/101/202 data split throughout. More details on the experimental parameters are available at our GitHub\footnote{\scriptsize \url{https://github.com/avishadilhara/Cross-Temporal-Sinhala-OCR}} repository.

\begin{table}[t]
\centering
\caption{Summary of all 8 fine-tuning experiments.}
\label{tab:experiments}
\footnotesize
\resizebox{0.85\columnwidth}{!}{%
\begin{tabular}{clllccc}
\hline
\textbf{Exp} & \textbf{Model} & \textbf{GPU} &
\textbf{Quant.} & \textbf{$r$} & \textbf{$\alpha$} &
\textbf{Dropout} \\
\hline
1 & DeepSeek-OCR V1  & P100 16\,GB   & 4-bit NF4 & 16 & 16 & 0 \\
2 & DeepSeek-OCR V2  & A100 80\,GB   & 16-bit    & 32 & 64 & 0.1 \\
3 & DeepSeek-OCR V2  & A100 80\,GB   & 16-bit    & 16 & 16 & 0 \\
4 & DeepSeek-OCR V2  & RTX 5090      & 4-bit NF4 & 16 & 16 & 0 \\
5 & LightOnOCR-2-1B  & P100 16\,GB   & 4-bit NF4 & 16 & 16 & 0.05 \\
6 & LightOnOCR-2-1B  & P100 16\,GB   & 4-bit NF4 & 32 & 64 & 0.1 \\
7 & LightOnOCR-2-1B  & RTX 4090      & 4-bit NF4 & 32 & 64 & 0.1 \\
8 & DeepSeek-OCR V2  & RTX 3090      & 4-bit NF4 & 16 & 16 & 0 \\
\hline
\end{tabular}}
\end{table}

\subsection{Evaluation Metrics}

All 8 fine-tuned models and baselines were evaluated on the same 202-sample held-out test set using five metrics: \textbf{CER} (Character Error Rate) and \textbf{WER} (Word Error Rate), computed via edit distance at the character and word levels
respectively (lower is better); and \textbf{BLEU}, \textbf{METEOR}, and \textbf{ANLS} (Average Normalised Levenshtein Similarity), measuring n-gram precision, fluency, and string similarity respectively (higher is better). 
script-level OCR quality.

\section{Results}

\subsection{Pre-Trained Baseline Performance}

Before fine-tuning, the three VLMs were evaluated on the test dataset (202 samples) in a zero-shot manner. None of the models performed satisfactorily; DeepSeek-OCR~V1 generated a CER of 61.46\%. However, DeepSeek-OCR~V2 (96.11\%) and LightOnOCR-2-1B (88.05\%) failed almost completely, which shows that zero-shot Sinhala OCR using present-day VLMs is not possible with current VLMs due to insufficient pre-training coverage \cite{jayatilleke_zero-shot_2025, vasantharajan_adapting_2022}. Table~\ref{tab:baseline} reports the full zero-shot metrics.

\begin{table}[t]
\centering
\caption{Pre-trained zero-shot performance before fine-tuning}
\label{tab:baseline}
\footnotesize
\resizebox{0.85\columnwidth}{!}{%
\begin{tabular}{clccccc}
\hline
\textbf{Exp} & \textbf{Model} & \textbf{CER↓} & \textbf{WER↓}
& \textbf{METEOR↑} & \textbf{BLEU↑} & \textbf{ANLS↑} \\
\hline
1 & DeepSeek-OCR V1  & 0.6146 & 0.7817 & 0.3754 & 0.3749 & 0.3854 \\
2 & DeepSeek-OCR V2  & 0.9611 & 0.9930 & 0.1940 & 0.0585 & 0.0389 \\
3 & LightOnOCR-2-1B  & 0.8805 & 0.9896 & 0.1311 & 0.0316 & 0.1195 \\
\hline
\end{tabular}}
\end{table}

\subsection{Fine-Tuning Results}

After fine-tuning with QLoRA across 8 experiments, all models
showed dramatic improvements over their baselines.
Table~\ref{tab:finetune} summarises the performance of all 8
fine-tuned models on the same 202-sample test set.

\begin{table}[t]
\centering
\caption{Fine-tuned model performance after QLoRA adaptation (202 test samples).}
\label{tab:finetune}
\resizebox{0.85\columnwidth}{!}{%
\begin{tabular}{clccccc}
\hline
\textbf{Exp} & \textbf{Model (Config)} & \textbf{CER↓}
& \textbf{WER↓} & \textbf{METEOR↑} & \textbf{BLEU↑}
& \textbf{ANLS↑} \\
\hline
1 & DeepSeek-OCR V1      & 0.0302 & 0.1227 & 0.8441 & 0.9531 & 0.9698 \\
2 & DeepSeek-OCR V2      & 0.3471 & 0.4745 & 0.5191 & 0.7288 & 0.6529 \\
3 & DeepSeek-OCR V2      & 0.0694 & 0.1723 & 0.7739 & 0.9153 & 0.9306 \\
4 & DeepSeek-OCR V2      & 0.0694 & 0.1828 & 0.7794 & 0.9181 & 0.9306 \\
5 & LightOnOCR-2-1B      & 0.2517 & 0.3831 & 0.5342 & 0.7329 & 0.7483 \\
6 & LightOnOCR-2-1B      & 0.1413 & 0.2118 & 0.6541 & 0.8332 & 0.8587 \\
7 & LightOnOCR-2-1B (1540px) & \textbf{0.0105} & \textbf{0.0563} & \textbf{0.9492} & \textbf{0.9808} & \textbf{0.9895} \\
8 & DeepSeek-OCR V2 (r16, RTX3090)    & 0.0831 & 0.1975 & 0.7602 & 0.9046 & 0.9169 \\
\hline
\multicolumn{7}{l}{\textbf{Bold} = best overall.}
\end{tabular}}
\end{table}

In Experiment~7, we demonstrated LightOnOCR-2-1B\footnote{\scriptsize
\url{https://huggingface.co/avishadilhara/sinhala-lightonocr-2-1b-Qlora}}, which achieved the lowest CER (1.05\%) and WER (5.63\%). The reason for the success of the longest edge of 1540\,px compared to 700\,px was the Input resolution, which reduced CER from 25.17\% to 1.05\% and enabled the model to identify finer Sinhala stroke diacritics. In contrast, when analysing the DeepSeek-OCR V2 experiments, Experiment 2\footnote{\scriptsize \url{https://huggingface.co/avishadilhara/sinhala-deepseek-ocr-Qlora}}, which used r = 32 and a dropout rate of 0.1, produced significantly better results than Experiments 3 and 4, which used r = 16 and no dropout. This finding confirms that aggressive regularisation can be counterproductive when working with small datasets \cite{baek_what_2021}.

\subsection{Benchmarking Against Existing OCR Engines}

The three best-fine tuned models (Experiments~7, 1, and 3) were compared against four existing OCR Engines in the same 202-sample real-world test set: Google Document~AI, Surya-OCR \cite{jayatilleke_zero-shot_2025}, Tesseract~v5 \cite{vasantharajan_adapting_2022}, and Subasa-OCR, an open-source Sinhala-specific engine, in which the results are summarised in Table~\ref{tab:benchmark}.

\begin{table}[t]
\centering
\caption{Benchmarking fine-tuned models vs.\ existing OCR engines
on the \texttt{sinhala-ocr-lk-acts-1010} test set (202 samples).}
\label{tab:benchmark}
\resizebox{0.85\columnwidth}{!}{%
\begin{tabular}{clccccc}
\hline
\textbf{Rank} & \textbf{Model} & \textbf{CER↓} & \textbf{WER↓}
& \textbf{METEOR↑} & \textbf{BLEU↑} & \textbf{ANLS↑} \\
\hline
1 & LightOnOCR-2-1B (Exp~7) & \textbf{0.0105} & \textbf{0.0563} & 0.9492 & \textbf{0.9808} & \textbf{0.9895} \\
2 & Google Document AI       & 0.0206 & 0.0826 & \textbf{0.9609} & 0.9724 & 0.9797 \\
3 & DeepSeek-OCR V1 (Exp~1) & 0.0302 & 0.1227 & 0.8441 & 0.9531 & 0.9698 \\
4 & DeepSeek-OCR V2 (Exp~3) & 0.0694 & 0.1723 & 0.7739 & 0.9153 & 0.9306 \\
5 & Surya-OCR            & 0.0884 & 0.2664 & 0.7806 & 0.8877 & 0.9116 \\
6 & Subasa-OCR            & 0.0999 & 0.5251 & 0.7483 & 0.8355 & 0.9006 \\
7 & Tesseract v5          & 0.1069 & 0.4514 & 0.7947 & 0.8117 & 0.8951 \\
\hline
\textbf{Bold} = best overall.
\end{tabular}}
\end{table}

LightOnOCR-2-1B (Exp~7) performed the best with CER of 1.05\% and WER of 5.63\%, surpassing Google Document~AI (CER~2.06\%), a paid commercial system without available model weights. Surya-OCR served as the optimal open-source baseline, scoring CER~8.84\% and WER~26.64\%, although WER was much higher than that of the two best-performing models. The score achieved by Surya for synthetic Sinhala images in the study by Jayatilleke and de Silva \cite{jayatilleke_zero-shot_2025} shows a marked difference compared to the OCR accuracy results for our degraded real-world documents.

\section{Diachronic Analysis}

\subsection{Temporal Distribution of the Test Set}

The test set spans three print periods: \textbf{1981--1989} ($n{=}82$, high degradation ink fading, bleed-through, and noisy); \textbf{2000--2009} ($n{=}54$,
moderate degradation, desktop publishing era with minor defects); and \textbf{2010--2019} ($n{=}66$, modern digitally produced with minimal degradation) \cite{springmann_ocr_2017, anuradha_estimating_2021, schultze_chronicling_2025}.

\subsection{Period-Level Aggregated Performance}

Table~\ref{tab:diachronic_cer_wer} reports the sample-weighted mean CER and WER for all six models across three periods. Period-level scores are computed as sample-weighted means across individual publication years to account for unequal representation per year.

\begin{table}[t]
\centering
\caption{Period-level weighted mean CER (\%) and WER (\%) for fine-tuned models and baselines.}
\label{tab:diachronic_cer_wer}
\resizebox{0.85\columnwidth}{!}{%
\begin{tabular}{lcccccc}
\hline
\multirow{2}{*}{\textbf{Model}} & \multicolumn{2}{c}{\textbf{1981--89 (High)}}
& \multicolumn{2}{c}{\textbf{2000--09 (Mod.)}}
& \multicolumn{2}{c}{\textbf{2010--19 (Modern)}} \\
 & CER & WER & CER & WER & CER & WER \\
\hline
LightOnOCR-2-1B (Exp~7) & \textbf{1.66} & \textbf{8.70} & \textbf{0.71} & 4.42 & \textbf{0.58} & 2.81 \\
DeepSeek-OCR V1 (Exp~1) & 4.65 & 17.62 & 2.35 & 9.64 & 1.54 & 7.78 \\
Google Document AI       & 3.26 & 15.86 & 1.30 & \textbf{4.36} & 1.18 & \textbf{2.01} \\
Surya-OCR                & 12.80 & 37.40 & 7.35 & 22.24 & 5.15 & 16.85 \\
Tesseract v5             & 18.07 & 69.86 & 7.62 & 32.64 & 4.05 & 24.66 \\
Subasa-OCR               & 14.82 & 67.19 & 8.07 & 46.01 & 5.54 & 39.60 \\
\hline
\multicolumn{7}{l}{\textbf{Bold} = best per period per metric.}
\end{tabular}}
\end{table}

\begin{figure}[t]
    \centering
    \includegraphics[width=0.88 \linewidth]{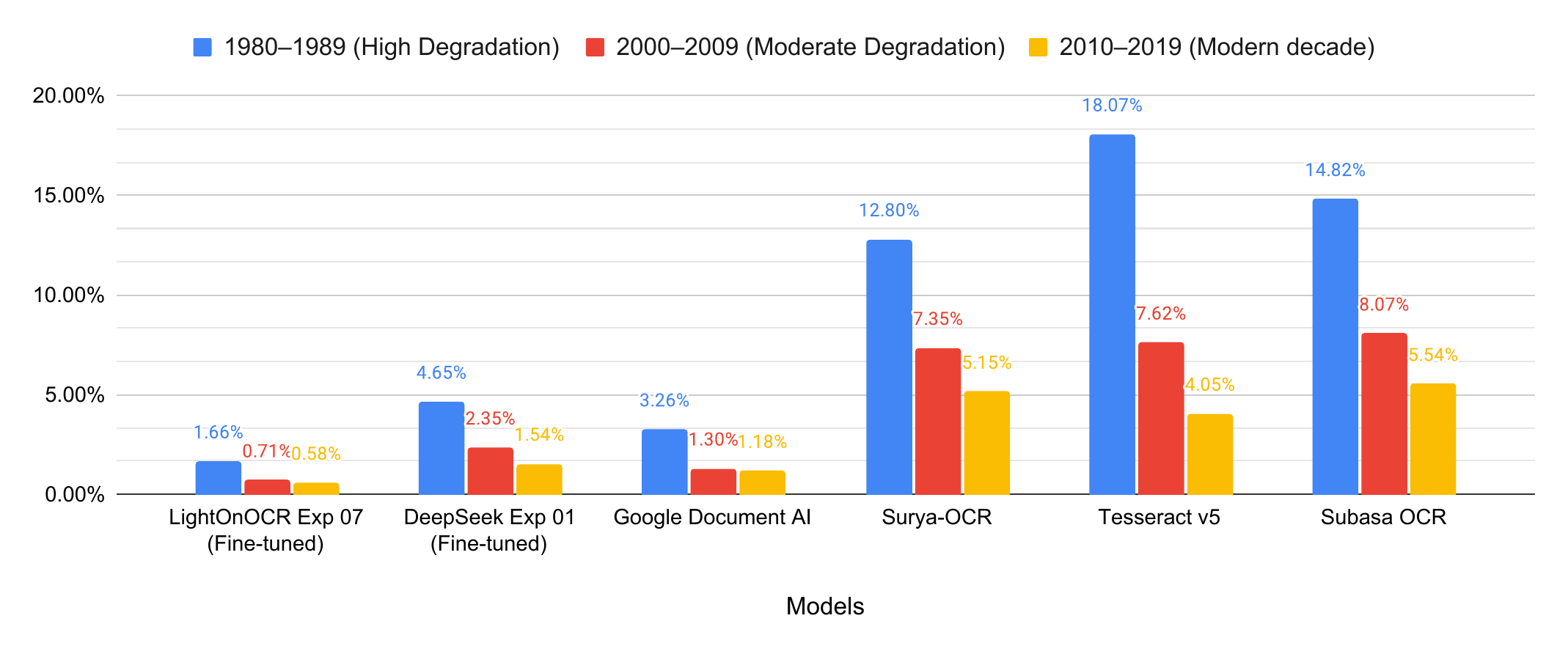}
    \caption{Sample-weighted mean CER across three temporal periods
    for all six models.}
    \label{fig:cer_by_period}
\end{figure}

\begin{figure}[t]
    \centering
    \includegraphics[width=0.88 \linewidth]{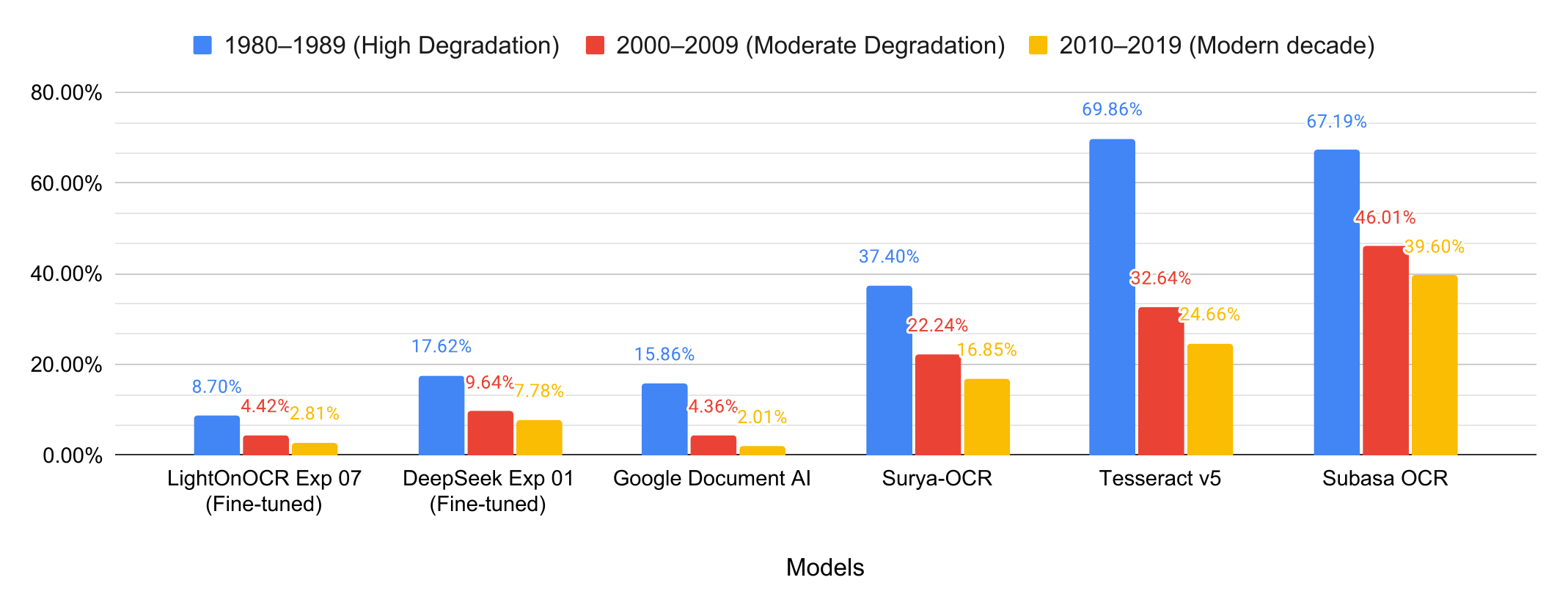}
    \caption{Sample-weighted mean WER across three temporal periods}
    \label{fig:wer_by_period}
\end{figure}

In all cases, there is a clear diachronic decline pattern in CER and WER, with values increasing as documents become older and more physically degraded (Figs.~\ref{fig:cer_by_period} and~\ref{fig:wer_by_period}). This result supports the hypothesis put forth by \cite{springmann_ocr_2017, schultze_chronicling_2025}, which suggests that OCR accuracy consistently declines as historical print quality worsens. In this study, we offer an analysis that confirms this theory for the Sinhala language, examining it diachronically for the first time.

\subsection{Legacy Baselines: Diachronic Collapse}

The findings from the diachronic analysis highlight a key shortcoming of traditional OCR tools in handling realistic historical Sinhala documents. In this regard, the OCR tools Tesseract~v5 and Subasa-OCR achieve WERs of 69.86\% and 67.19\% respectively on the 1980s documents, suggesting that around seven out of ten words are misrecognised. Therefore, line-segmentation-based OCR systems, which are built for working with clean binary images, are essentially incapable of handling degraded historical Sinhala documents \cite{vempati_why_2025}. Although Surya-OCR is currently the best-performing open-source baseline with a WER of 37.40\% in the 1980s, its performance is significantly inferior compared to the reported WER of 2.61\% on simulated documents \cite{jayatilleke_zero-shot_2025} a gap, confirming that synthetic benchmarks substantially overestimate real-world performance.

\subsection{BLEU, METEOR, and ANLS Across Periods}

Table~\ref{tab:diachronic_bleu} reports BLEU, METEOR, and ANLS across the three periods, confirming that the trends observed in CER/WER are consistent across all five evaluation metrics. LightOnOCR-2-1B achieves BLEU scores of 97.15\%, 98.50\%, and 98.90\% across the three periods, indicating that high n-gram precision is maintained even on degraded 1980s pages.

\begin{table}[t]
\centering
\caption{Period-level weighted mean BLEU, METEOR, and ANLS for
fine-tuned models and baselines.}
\label{tab:diachronic_bleu}
\resizebox{0.85\columnwidth}{!}{%
\begin{tabular}{lcccccccccc}
\hline
& \multicolumn{3}{c}{\textbf{1981--89}}
& \multicolumn{3}{c}{\textbf{2000--09}}
& \multicolumn{3}{c}{\textbf{2010--19}} \\
\textbf{Model} & BLEU & MTR & ANLS & BLEU & MTR & ANLS
& BLEU & MTR & ANLS \\
\hline
LightOnOCR (Exp 7) & \textbf{97.15} & 91.11 & \textbf{98.34} & \textbf{98.50} & 97.52 & \textbf{99.29} & \textbf{98.90} & 97.54 & \textbf{99.42} \\
DeepSeek V1 (Exp 1) & 93.17 & 75.97 & 95.35 & 96.13 & 89.27 & 97.65 & 97.31 & 90.92 & 98.46 \\
Google Doc AI       & 95.65 & \textbf{92.25} & 96.79 & 98.17 & \textbf{97.91} & 98.71 & 98.47 & \textbf{99.38} & 98.84 \\
Surya-OCR           & 84.01 & 66.14 & 87.20 & 90.60 & 82.37 & 92.65 & 93.18 & 89.34 & 94.85 \\
Tesseract v5        & 70.01 & 58.99 & 83.30 & 85.98 & 87.66 & 90.93 & 91.10 & 98.21 & 96.07 \\
Subasa-OCR          & 74.87 & 57.35 & 85.26 & 87.56 & 81.43 & 91.94 & 91.05 & 91.16 & 94.47 \\
\hline
\multicolumn{10}{l}{MTR = METEOR. All values in \%.} \\
\multicolumn{10}{l}{\textbf{Bold} = best per period per metric.} \\
\end{tabular}}
\end{table}

\section{Discussion}

\subsection{Fine-Tuning, and Commercial Comparison}

With only 707 training data points in the real world, QLoRA-based fine-tuning enabled all three VLMs to move from being unable to recognise Sinhala words to achieving high accuracy. LightOnOCR-2-1B lowered its CER from 88.05\% to 1.05\% (with an overall improvement of 86.97\%), whereas DeepSeek-OCR~V1 went from 61.46\% to 3.02\%, following the same pattern as other work showing that PEFT based on LoRA requires far fewer samples than full fine-tuning for low-resource Indic scripts \cite{kolavi_nayana_2025, agarwal_concise_2024}. 

Fine-tuned LightOnOCR-2-1B (CER~1.05\%) outperformed Google Document~AI (CER~2.06\%) according to the main measure, but Document~AI had a slight advantage on METEOR (96.09\% compared to 94.92\%). This analysis is clear: Document~AI output was solely used as an initial annotation seed and manually corrected character by character before being used as ground truth, avoiding circularity in the evaluation.

\subsection{Diachronic Degradation and Dataset Significance}

Diachronic degradation is a common indicator for all six systems: LightOnOCR-2-1B CER increases by 2.9$\times$ from the contemporary period (0.58\%) to the highly degraded period of 1981--1989 (1.66\%), whereas Tesseract~v5 increases by 4.5$\times$ (4.05\% $\to$ 18.07\%). Subasa-OCR attains WER~67.19\% on texts from the 1980s, demonstrating that almost seven out of ten words are recognized incorrectly \cite{springmann_ocr_2017}. Notably, the reduced diachronic sensitivity of these fine-tuned VLMs can be attributed to their domain-specific training using actual degraded data, and not any inherent structural superiority over line segmentation models \cite{vasantharajan_adapting_2022, baek_what_2021}. The performance of a fine-tuned line-level model in achieving similar resilience remains an open question.

\subsection{Limitations}

There are several limitations that constrain the validity of these findings and should be addressed in future work.

\vspace{0.4em}
\noindent
\textbf{Domain homogeneity:}
The corpus contains only Sri Lankan legislative statutes. Although
this guarantees consistency, it limits the diversity of fonts, layout, and vocabulary. The performance could vary widely across Sinhala in newspapers, textbooks, and handwritten Sinhala.

\vspace{0.4em}
\noindent
\textbf{Temporal coverage gap:}
The decade from 1990 to 1999 was excluded due to the infeasibility of manual annotation within the project timeframe. 

\vspace{0.4em}
\noindent
\textbf{Year-level sample imbalance:}
There is a large variation in the number of samples per year within each time period; for example, 1988 has 26 pages available, whereas some years have only 1 to 2 pages. Time-period-level weighted averages solve this problem, yet year-based point averages remain statistically unreliable.

\section{Conclusion}

The fine-tuned LightOnOCR-2-1B (Exp~7) model obtained a CER of \textbf{1.05\%} and a WER of \textbf{5.63\%} on the realistic test set, outperforming all the open-source baselines and Google Document~AI (CER~2.06\%) models, proving that a small 1B-parameter open-source VLM optimised on 707 realistic samples can outperform a commercial OCR engine in this setting. For the first time, a diachronic evaluation of Sinhala OCR was conducted, showing that the degradation of documents during printing periods is a universal factor influencing performance in all six models, increasing the CER by a factor of 2.9$\times$ in the case of LightOnOCR-2-1B and 4.5$\times$ in the case of Tesseract~v5 when comparing the modern and highly degraded periods. Importantly, the decreased sensitivity of the optimised models to diachronic degradation cannot be explained by architectural differences but rather by domain-specific optimisation on degraded examples, which remains a question for future experimental studies.

Further research can include expanding the corpus to include additional types of Sinhala documents (newspapers, textbooks, gazettes), the missing decade 1990-1999, and conducting a diachronic comparison using controlled architectural experiments with fine-tuned line-level and page-level models.

{\footnotesize
\bibliographystyle{IEEEtran}
\bibliography{My_research_docs}

@inproceedings{kolavi_nayana_2025,
    title = "Nayana {OCR}: A Scalable Framework for Document {OCR} in Low-Resource Languages",
    author = "Kolavi, Adithya  and
      P, Samarth  and
      Jain, Vyoman",
    editor = "Truong, Sang  and
      Putri, Rifki Afina  and
      Nguyen, Duc  and
      Wang, Angelina  and
      Ho, Daniel  and
      Oh, Alice  and
      Koyejo, Sanmi",
    booktitle = "Proceedings of the 1st Workshop on Language Models for Underserved Communities (LM4UC 2025)",
    month = may,
    year = "2025",
    address = "Albuquerque, New Mexico",
    publisher = "Association for Computational Linguistics",
    url = "https://aclanthology.org/2025.lm4uc-1.11/",
    doi = "10.18653/v1/2025.lm4uc-1.11",
    pages = "86--103",
    ISBN = "979-8-89176-242-8"
}

@article{anuradha_estimating_2021,
	title = {Estimating the Effects of Text Genre, Image Resolution and Algorithmic Complexity needed for Sinhala Optical Character Recognition},
	volume = {14},
	issn = {2550-2794, 1800-4156},
	url = {https://account.icter.sljol.info/index.php/sljo-j-ijaicterict/article/view/7231},
	doi = {10.4038/icter.v14i3.7231},
	pages = {43--51},
	number = {3},
	journaltitle = {Int J on Adv. in {ICT} for Emerging Countries},
	author = {Anuradha, Isuri and Liyanage, Chamila and Weerasinghe, Ruvan},
	urldate = {2025-08-06},
	year = {2021},
	langid = {english},
}

@article{purushoth_velayuthan_benchmarking_2025,
	title = {Benchmarking {OCR} Models for Sinhala and Tamil Document Digitization},
	url = {https://rgdoi.net/10.13140/RG.2.2.20843.25129},
	doi = {10.13140/RG.2.2.20843.25129},
	publisher = {Unpublished},
	author = {{Purushoth Velayuthan} and {Thanuja D Ambegoda}},
	urldate = {2025-08-06},
	year = {2025},
	langid = {english},
}

@inproceedings{vasantharajan_adapting_2022,
	location = {Singapore, Singapore},
	title = {Adapting the Tesseract Open-Source {OCR} Engine for Tamil and Sinhala Legacy Fonts and Creating a Parallel Corpus for Tamil-Sinhala-English},
	rights = {https://doi.org/10.15223/policy-029},
	isbn = {978-1-6654-7674-4},
	url = {https://ieeexplore.ieee.org/document/9961304/},
	doi = {10.1109/IALP57159.2022.9961304},
	eventtitle = {2022 International Conference on Asian Language Processing ({IALP})},
	pages = {143--149},
	booktitle = {2022 International Conference on Asian Language Processing ({IALP})},
	publisher = {{IEEE}},
	author = {Vasantharajan, Charangan and Tharmalingam, Laksika and Thayasivam, Uthayasanker},
	urldate = {2025-08-04},
	year = {2022},
	langid = {english},
}

@inproceedings{anuradha_deep_2020,
	location = {Colombo, Sri Lanka},
	title = {Deep Learning Based Sinhala Optical Character Recognition ({OCR})},
	rights = {https://ieeexplore.ieee.org/Xplorehelp/downloads/license-information/{IEEE}.html},
	isbn = {978-1-7281-8655-9},
	url = {https://ieeexplore.ieee.org/document/9325428/},
	doi = {10.1109/ICTer51097.2020.9325428},
	eventtitle = {2020 20th International Conference on Advances in {ICT} for Emerging Regions ({ICTer})},
	pages = {298--299},
	booktitle = {2020 20th International Conference on Advances in {ICT} for Emerging Regions ({ICTer})},
	publisher = {{IEEE}},
	author = {Anuradha, Isuri and Liyanage, Chamila and Wijayawardhana, Harsha and Weerasinghe, Ruvan},
	urldate = {2025-08-04},
	year = {2020},
	langid = {english},
}

@article{shi_aster_2019,
	title = {{ASTER}: An Attentional Scene Text Recognizer with Flexible Rectification},
	volume = {41},
	rights = {https://ieeexplore.ieee.org/Xplorehelp/downloads/license-information/{IEEE}.html},
	issn = {0162-8828, 2160-9292, 1939-3539},
	url = {https://ieeexplore.ieee.org/document/8395027/},
	doi = {10.1109/TPAMI.2018.2848939},
	shorttitle = {{ASTER}},
	pages = {2035--2048},
	number = {9},
	journaltitle = {{IEEE} Trans. Pattern Anal. Mach. Intell.},
	author = {Shi, Baoguang and Yang, Mingkun and Wang, Xinggang and Lyu, Pengyuan and Yao, Cong and Bai, Xiang},
	urldate = {2025-07-28},
	year = {2019},
	langid = {english},
}

@inproceedings{baek_character_2019,
	location = {Long Beach, {CA}, {USA}},
	title = {Character Region Awareness for Text Detection},
	rights = {https://ieeexplore.ieee.org/Xplorehelp/downloads/license-information/{IEEE}.html},
	url = {https://ieeexplore.ieee.org/document/8953846/},
	doi = {10.1109/cvpr.2019.00959},
	eventtitle = {2019 {IEEE}/{CVF} Conference on Computer Vision and Pattern Recognition ({CVPR})},
	pages = {9357--9366},
	booktitle = {2019 {IEEE}/{CVF} Conference on Computer Vision and Pattern Recognition ({CVPR})},
	publisher = {{IEEE}},
	author = {Baek, Youngmin and Lee, Bado and Han, Dongyoon and Yun, Sangdoo and Lee, Hwalsuk},
	urldate = {2025-07-24},
	year = {2019},
	langid = {english},
}

@inproceedings{baek_what_2021,
	location = {Nashville, {TN}, {USA}},
	title = {What If We Only Use Real Datasets for Scene Text Recognition? Toward Scene Text Recognition With Fewer Labels},
	rights = {https://doi.org/10.15223/policy-029},
	url = {https://ieeexplore.ieee.org/document/9578847/},
	doi = {10.1109/cvpr46437.2021.00313},
	shorttitle = {What If We Only Use Real Datasets for Scene Text Recognition?},
	eventtitle = {2021 {IEEE}/{CVF} Conference on Computer Vision and Pattern Recognition ({CVPR})},
	booktitle = {2021 {IEEE}/{CVF} Conference on Computer Vision and Pattern Recognition ({CVPR})},
	publisher = {{IEEE}},
	author = {Baek, Jeonghun and Matsui, Yusuke and Aizawa, Kiyoharu},
	urldate = {2025-07-24},
	year = {2021},
	langid = {english},
}

@inproceedings{shi_robust_2016,
  title={Robust scene text recognition with automatic rectification},
  author={Shi, Baoguang and Wang, Xinggang and Lyu, Pengyuan and Yao, Cong and Bai, Xiang},
  booktitle={Proceedings of the IEEE conference on computer vision and pattern recognition},
  pages={4168--4176},
  year={2016}
}

@inproceedings{zhou_east_2017,
  title={East: an efficient and accurate scene text detector},
  author={Zhou, Xinyu and Yao, Cong and Wen, He and Wang, Yuzhi and Zhou, Shuchang and He, Weiran and Liang, Jiajun},
  booktitle={Proceedings of the IEEE conference on Computer Vision and Pattern Recognition},
  pages={5551--5560},
  year={2017}
}

@article{shi_end--end_2017,
  title={An end-to-end trainable neural network for image-based sequence recognition and its application to scene text recognition},
  author={Shi, Baoguang and Bai, Xiang and Yao, Cong},
  journal={IEEE transactions on pattern analysis and machine intelligence},
  volume={39},
  number={11},
  pages={2298--2304},
  year={2016},
  publisher={IEEE}
}

@article{dosovitskiy_image_2021,
	title = {An Image is Worth 16x16 Words: Transformers for Image Recognition at Scale},
	author = {Dosovitskiy, Alexey and Beyer, Lucas and Kolesnikov, Alexander and Weissenborn, Dirk and Zhai, Xiaohua and Unterthiner, Thomas and Dehghani, Mostafa and Minderer, Matthias and Heigold, Georg and Gelly, Sylvain and Uszkoreit, Jakob and Houlsby, Neil},
	year = {2021},
	langid = {english},
}

@article{vaswani_attention_nodate,
  title={Attention is all you need},
  author={Vaswani, Ashish and Shazeer, Noam and Parmar, Niki and Uszkoreit, Jakob and Jones, Llion and Gomez, Aidan N and Kaiser, {\L}ukasz and Polosukhin, Illia},
  journal={Advances in neural information processing systems},
  volume={30},
  year={2017}
}

@article{li_trocr_2023,
	title = {{TrOCR}: Transformer-Based Optical Character Recognition with Pre-trained Models},
	volume = {37},
	issn = {2374-3468, 2159-5399},
	url = {https://ojs.aaai.org/index.php/AAAI/article/view/26538},
	doi = {10.1609/aaai.v37i11.26538},
	shorttitle = {{TrOCR}},
	pages = {13094--13102},
	number = {11},
	journaltitle = {{AAAI}},
	publisher = {Association for the Advancement of Artificial Intelligence ({AAAI})},
	author = {Li, Minghao and Lv, Tengchao and Chen, Jingye and Cui, Lei and Lu, Yijuan and Florencio, Dinei and Zhang, Cha and Li, Zhoujun and Wei, Furu},
	urldate = {2025-07-16},
	year = {2023},
	langid = {english},
}

@inproceedings{bissacco_photoocr_2013,
	location = {Sydney, Australia},
	title = {{PhotoOCR}: Reading Text in Uncontrolled Conditions},
	url = {http://ieeexplore.ieee.org/document/6751207/},
	doi = {10.1109/iccv.2013.102},
	shorttitle = {{PhotoOCR}},
	eventtitle = {2013 {IEEE} International Conference on Computer Vision ({ICCV})},
	pages = {785--792},
	booktitle = {2013 {IEEE} International Conference on Computer Vision},
	publisher = {{IEEE}},
	author = {Bissacco, Alessandro and Cummins, Mark and Netzer, Yuval and Neven, Hartmut},
	urldate = {2025-07-14},
	year = {2013},
	langid = {english},
}

@inproceedings{agarwal_concise_2024,
	location = {Mexico City, Mexico},
	title = {A Concise Survey of {OCR} for Low-Resource Languages},
	url = {https://aclanthology.org/2024.americasnlp-1.10},
	doi = {10.18653/v1/2024.americasnlp-1.10},
	eventtitle = {Proceedings of the 4th Workshop on Natural Language Processing for Indigenous Languages of the Americas ({AmericasNLP} 2024)},
	pages = {88--102},
	booktitle = {Proceedings of the 4th Workshop on Natural Language Processing for Indigenous Languages of the Americas ({AmericasNLP} 2024)},
	publisher = {Association for Computational Linguistics},
	author = {Agarwal, Milind and Anastasopoulos, Antonios},
	urldate = {2025-08-16},
	year = {2024},
	langid = {english},
}

@article{hegghammer_ocr_2021,
	title = {{OCR} with Tesseract, Amazon Textract, and Google Document {AI}: a benchmarking experiment},
	volume = {5},
	issn = {2432-2717, 2432-2725},
	url = {https://link.springer.com/10.1007/s42001-021-00149-1},
	doi = {10.1007/s42001-021-00149-1},
	shorttitle = {{OCR} with Tesseract, Amazon Textract, and Google Document {AI}},
	pages = {861--882},
	number = {1},
	journaltitle = {J Comput Soc Sc},
	author = {Hegghammer, Thomas},
	urldate = {2025-08-17},
	year = {2021},
	langid = {english},
}

@inproceedings{jayatilleke_zero-shot_2025,
    title = "Zero-shot {OCR} Accuracy of Low-Resourced Languages: A Comparative Analysis on {S}inhala and {T}amil",
    author = "Jayatilleke, Nevidu  and
      de Silva, Nisansa",
    editor = "Angelova, Galia  and
      Kunilovskaya, Maria  and
      Escribe, Marie  and
      Mitkov, Ruslan",
    booktitle = "Proceedings of the 15th International Conference on Recent Advances in Natural Language Processing - Natural Language Processing in the Generative AI Era",
    month = sep,
    year = "2025",
    address = "Varna, Bulgaria",
    publisher = "INCOMA Ltd., Shoumen, Bulgaria",
    url = "https://aclanthology.org/2025.ranlp-1.56/",
    pages = "471--480"
}

@misc{wei_deepseek-ocr_2025,
	title = {{DeepSeek}-{OCR}: Contexts Optical Compression},
	url = {http://arxiv.org/abs/2510.18234},
	doi = {10.48550/arXiv.2510.18234},
	shorttitle = {{DeepSeek}-{OCR}},
	number = {{arXiv}:2510.18234},
	publisher = {{arXiv}},
	author = {Wei, Haoran and Sun, Yaofeng and Li, Yukun},
	urldate = {2025-10-31},
	year = {2025},
	langid = {english},
	eprinttype = {arxiv},
	eprint = {2510.18234 [cs]},
}

@misc{senaratna_sri_2025,
	title = {Sri Lanka Document Datasets: A Large-Scale, Multilingual Resource for Law, News, and Policy},
	url = {http://arxiv.org/abs/2510.04124},
	doi = {10.48550/arXiv.2510.04124},
	shorttitle = {Sri Lanka Document Datasets},
	number = {{arXiv}:2510.04124},
	publisher = {{arXiv}},
	author = {Senaratna, Nuwan I.},
	urldate = {2025-11-05},
	year = {2025},
	langid = {english},
	eprinttype = {arxiv},
	eprint = {2510.04124 [cs]},
}

@article{dettmers_qlora_2023,
	title = {{QLORA}: Efficient Finetuning of Quantized {LLMs}},
	author = {Dettmers, Tim and Pagnoni, Artidoro and Holtzman, Ari and Zettlemoyer, Luke},
	date = {2023},
	langid = {english},
}

@IEEEtranBSTCTL{IEEEexample:BSTcontrol,
  CTLdash_repeated_names = "no"
}

@article{wei_deepseek-ocr_2026,
  title={Deep{S}eek-{OCR} 2: Visual Causal Flow},
  author={Wei, Haoran and Sun, Yaofeng and Li, Yukun},
  journal={arXiv preprint arXiv:2601.20552},
  year={2026}
}

@misc{taghadouini_lightonocr_2026,
	title = {{LightOnOCR}: A 1B End-to-End Multilingual Vision-Language Model for State-of-the-Art {OCR}},
	url = {http://arxiv.org/abs/2601.14251},
	doi = {10.48550/arXiv.2601.14251},
	shorttitle = {{LightOnOCR}},
	number = {{arXiv}:2601.14251},
	publisher = {{arXiv}},
	author = {Taghadouini, Said and Cavaillès, Adrien and Aubertin, Baptiste},
	urldate = {2026-03-07},
	year = {2026},
	langid = {english},
	eprinttype = {arxiv},
	eprint = {2601.14251 [cs]},
}

@misc{schultze_chronicling_2025,
	title = {Chronicling Germany: An Annotated Historical Newspaper Dataset},
	url = {http://arxiv.org/abs/2401.16845},
	doi = {10.48550/arXiv.2401.16845},
	shorttitle = {Chronicling Germany},
	number = {{arXiv}:2401.16845},
	publisher = {{arXiv}},
	author = {Schultze, Christian and Kerkfeld, Niklas and Kuebart, Kara and Weber, Princilia and Wolter, Moritz and Selgert, Felix},
	urldate = {2026-03-25},
	year = {2025},
	langid = {english},
	eprinttype = {arxiv},
	eprint = {2401.16845 [cs]},
}

@misc{springmann_ocr_2017,
	title = {{OCR} of historical printings with an application to building diachronic corpora: A case study using the {RIDGES} herbal corpus},
	url = {http://arxiv.org/abs/1608.02153},
	doi = {10.48550/arXiv.1608.02153},
	shorttitle = {{OCR} of historical printings with an application to building diachronic corpora},
	number = {{arXiv}:1608.02153},
	publisher = {{arXiv}},
	author = {Springmann, U. and Lüdeling, A.},
	urldate = {2026-03-25},
	year = {2017},
	langid = {english},
	eprinttype = {arxiv},
	eprint = {1608.02153 [cs]},
}

@misc{lv_kosmos-25_2024,
	title = {{KOSMOS}-2.5: A Multimodal Literate Model},
	url = {http://arxiv.org/abs/2309.11419},
	doi = {10.48550/arXiv.2309.11419},
	shorttitle = {{KOSMOS}-2.5},
	number = {{arXiv}:2309.11419},
	publisher = {{arXiv}},
	author = {Lv, Tengchao and Huang, Yupan and Chen, Jingye and Zhao, Yuzhong and Jia, Yilin and Cui, Lei and Ma, Shuming and Chang, Yaoyao and Huang, Shaohan and Wang, Wenhui and Dong, Li and Luo, Weiyao and Wu, Shaoxiang and Wang, Guoxin and Zhang, Cha and Wei, Furu},
	urldate = {2026-03-26},
	year = {2024},
	langid = {english},
	eprinttype = {arxiv},
	eprint = {2309.11419 [cs]},
}

@misc{yu_benchmarking_2025,
	title = {Benchmarking Vision-Language Models on Chinese Ancient Documents: From {OCR} to Knowledge Reasoning},
	url = {http://arxiv.org/abs/2509.09731},
	doi = {10.48550/arXiv.2509.09731},
	shorttitle = {Benchmarking Vision-Language Models on Chinese Ancient Documents},
	number = {{arXiv}:2509.09731},
	publisher = {{arXiv}},
	author = {Yu, Haiyang and Wu, Yuchuan and Shi, Fan and Liao, Lei and Lu, Jinghui and Ge, Xiaodong and Wang, Han and Zhuo, Minghan and Wu, Xuecheng and Fei, Xiang and Feng, Hao and Tang, Guozhi and Wang, An-Lan and Zhu, Hanshen and He, Yangfan and Liang, Quanhuan and Meng, Liyuan and Feng, Chao and Huang, Can and Tang, Jingqun and Li, Bin},
	urldate = {2026-03-26},
	year = {2025},
	langid = {english},
	eprinttype = {arxiv},
	eprint = {2509.09731 [cs]},
}

@misc{vempati_why_2025,
	title = {Why Stop at Words? Unveiling the Bigger Picture through Line-Level {OCR}},
	url = {http://arxiv.org/abs/2508.21693},
	doi = {10.48550/arXiv.2508.21693},
	shorttitle = {Why Stop at Words?},
	number = {{arXiv}:2508.21693},
	publisher = {{arXiv}},
	author = {Vempati, Shashank and Anand, Nishit and Talebailkar, Gaurav and Garai, Arpan and Arora, Chetan},
	urldate = {2026-03-26},
	year = {2025},
	langid = {english},
	eprinttype = {arxiv},
	eprint = {2508.21693 [cs]},
}

@misc{hu_lora_2021,
	title = {{LoRA}: Low-Rank Adaptation of Large Language Models},
	url = {http://arxiv.org/abs/2106.09685},
	doi = {10.48550/arXiv.2106.09685},
	shorttitle = {{LoRA}},
	number = {{arXiv}:2106.09685},
	publisher = {{arXiv}},
	author = {Hu, Edward J. and Shen, Yelong and Wallis, Phillip and Allen-Zhu, Zeyuan and Li, Yuanzhi and Wang, Shean and Wang, Lu and Chen, Weizhu},
	urldate = {2026-03-26},
	year = {2021},
	langid = {english},
	eprinttype = {arxiv},
	eprint = {2106.09685 [cs]},
}
}

\end{document}